\documentclass[final]{cvpr}

\usepackage{times}
\usepackage{epsfig}
\usepackage{graphicx}
\usepackage{amsmath}
\usepackage{amssymb}
\usepackage{subfigure}

\newcommand*\samethanks[1][\value{footnote}]{}
\renewcommand\footnotemark{}

\usepackage[pagebackref=true,breaklinks=true,colorlinks,bookmarks=false]{hyperref}

\def\cvprPaperID{0839} 
\def\confYear{CVPR 2021}

\begin{document}

\title{Online Exemplar Fine-Tuning for Image-to-Image Translation}

\author{
	Taewon Kang$^*$\thanks{$^*$Equal contribution}, 
	Soohyun Kim$^*$\samethanks, 
	Sunwoo Kim, and Seungryong Kim$^\dagger$\thanks{$^\dagger$Corresponding author}\\
	Computer Vision Lab (CVLAB), Korea University, Seoul, Korea\\ 
	\texttt{itschool@itsc.kr}, \texttt{lune1211@ewhain.net}, \texttt{seungryong\_kim@korea.ac.kr} \\
} 


\maketitle

\begin{abstract}
Existing techniques to solve exemplar-based image-to-image translation within deep convolutional neural networks (CNNs) generally require a training phase to optimize the network parameters on domain-specific and task-specific benchmarks, thus having limited applicability and generalization ability. In this paper, we propose a novel framework, for the first time, to solve exemplar-based translation through an online optimization given an input image pair, called online exemplar fine-tuning (OEFT), in which we fine-tune the off-the-shelf and general-purpose networks to the input image pair themselves. We design two sub-networks, namely correspondence fine-tuning and multiple GAN inversion, and optimize these network parameters and latent codes, starting from the pre-trained ones, with well-defined loss functions. Our framework does not require the off-line training phase, which has been the main challenge of existing methods, but the pre-trained networks to enable optimization in online. Experimental results prove that our framework is effective in having a generalization power to unseen image pairs and clearly even outperforms the state-of-the-arts needing the intensive training phase.
\end{abstract}


\section{Introduction}
For a decade, image-to-image translation (I2I), aiming at translating an image in one domain (i.e., source) to another domain (i.e., target), has been popularly studied and used in many applications, such as style transfer~\cite{gatys2016image,huang2017arbitrary}, super-resolution~\cite{dong2015image,kim2016accurate}, image inpainting~\cite{pathak2016context,iizuka2017globally}, and colorization~\cite{zhang2016colorful, zhang2017real}.
In particular, \emph{exemplar}-based image-to-image translation is a conditional image translation that reconstructs the image in source given a guidance of an exemplar in target, and has also achieved steady progress recently~\cite{huang2018multimodal, ma2018exemplar, wang2019example, zhang2020cross}.

\begin{figure}[t]
\begin{center}
\includegraphics[width=1\linewidth]{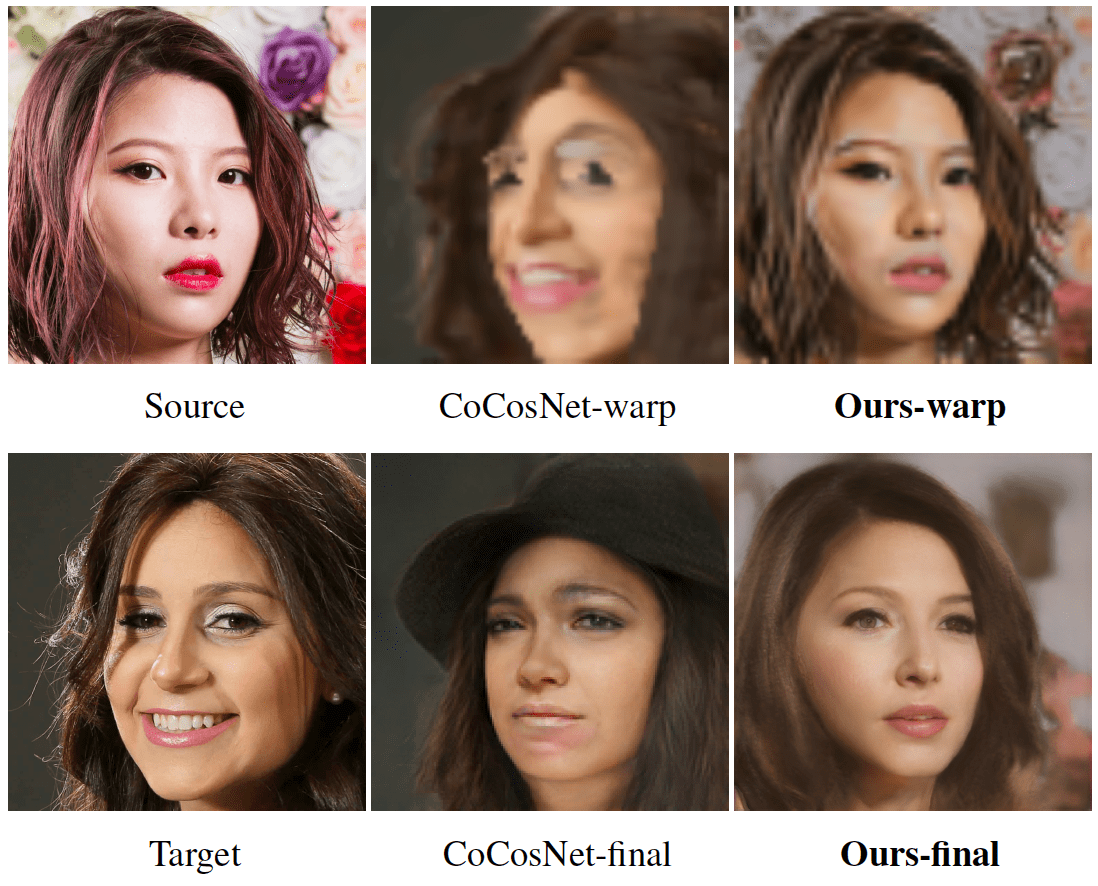}
\end{center}\vspace{-10pt}
  \caption{\textbf{Exemplar-based image-to-image translation results using our framework:} Existing methods, such as CoCosNet~\cite{zhang2020cross}, one of the state-of-the-arts, have shown a limited generalization ability to unseen input pairs, despite its intensive \textbf{\emph{training}} on large-scale dataset. Unlike this, our framework just deploys off-the-shelf feature detector and image generator, and \textbf{\emph{fine-tunes}} the networks on the input pair solely, having better generalization ability without training on specific image-to-image translation benchmarks.}
\label{exemplar_generation}\vspace{-10pt}
\end{figure}

Most recent methods directly \emph{learn} the mapping function from an image in one domain to output image in another domain by transferring the style of exemplar, through deep convolutional neural network (CNNs) due to its high capacity~\cite{gatys2016image,huang2017arbitrary,li2017universal}. In general, they first extract the style information from the exemplar in target domain globally~\cite{park2019semantic} or locally~\cite{zhu2020sean} and transfer it to the reference image in source domain, followed by the generation procedure~\cite{park2019semantic,zhu2020sean}. 
In literature~\cite{zhang2020cross,zhang2019deep}, the existing methods attempt to design their own \emph{correspondence} modules and \emph{translation} modules and train them on a large number of reference-exemplar samples as training supervision for each specific task they try to solve, e.g., mask-to-face, edge-to-face, day-to-night, and cat2dog. For instance, CoCosNet~\cite{zhang2020cross}, one of the recent state-of-the-art models, trains these networks and trains on each specific task with dataset. However, collecting such dataset is costly and labor-intensive, or even impossible. Moreover, it is a well-known fact that the network trained on specific images or tasks has limited generalization ability to unseen images. As exemplified in Figure~\ref{exemplar_generation}, \emph{pre-trained} CoCosNet~\cite{zhang2020cross} fails to translate unseen reference-exemplar pair. 
In addition, most of existing techniques for this task are based on generative adversarial networks (GANs), and it has been shown that training the GANs is notoriously challenging and requires high time-consumption and memory.

In this paper, we explore an alternative, simpler solution, called online exemplar fine-tuning (OEFT), to overcome aforementioned limitations for exemplar-based image-to-image translation, without need of any large-scale dataset, hand-labeling, and even task-specific training process. 
Our key-ingredient is to utilize an off-the-shelf network, which is already trained for general image classification~\cite{simonyan2014very,he2016deep} or image generation~\cite{goodfellow2014generative, karras2019style},  and \emph{fine-tune} the network to the input reference-exemplar pair to learn the mapping by considering internal statistics between them.
Such a strategy has been also proposed in other tasks~\cite{ulyanov2018deep,shaham2019singan, luo2017learning}, where they only access the pre-trained, general-purpose networks to solve their own problems, but not fully explored in exemplar-based image-to-image translation yet.   

Specifically, rather than directly finding such a mapping function, we present a two-stage solution consisting of correspondence fine-tuning(CFT) module and multiple GANs inversion (MGI) module. In the CFT, we predict a translation hypothesis by first measuring similarities between the point in source and other points in target, and then transferring the target image to the source, inspired by the classical matching pipeline~\cite{rocco2017convolutional}. Any off-the-shelf convolutional architecture to extract dense feature volumes in which we compute to get similarity can be used in this module, fine-tuned or optimized with the well-defined loss functions such as contrastive loss, perceptual loss, contextual loss, and regularization loss to extract the good representation considering the internal mapping relationship. To further refine the translation hypothesis, we also present a novel GANs inversion in the MGI in a manner that self-decides to find out the optimal translation result. 

Since our framework is not tailored to specific domain pairs or tasks, but a single \emph{input image pair}, it yields surprisingly outstanding generalization ability, as exemplified in Figure~\ref{exemplar_generation}, without any supervision or task-specific training. Our experiments on CelebA-HQ~\cite{liu2015deep} and Flickr Faces HQ~\cite{karras2019style} evidence that our framework consistently outperforms the existing methods.

Our contribution can be summarized as follows:
\begin{itemize}
	\item We propose online fine-tuning method for exemplar-based image-to-image translation, which does not require any task-specific domain dataset or training.
	\item We present correspondence fine-tuning module that establishes accurate correspondence fields by only considering the internal matching statistics.
	\item We propose multiple GANs inversion that selects more plausible image reconstruction result among multiple hypotheses without any training or supervision.
\end{itemize}

\section{Related Works}

\textbf{Exemplar-based Image-to-image translation.}
While most early efforts for image-to-Image (I2I) translation are based on supervision learning~\cite{isola2017image}, current state-of-the-arts focus \emph{unpaired} learning. For instance, CycleGAN~\cite{zhu2017unpaired} was the first attempt. After that, numerous other approaches have been presented and can be divided into unimodal translation~\cite{liu2017unsupervised,zhu2017unpaired,kim2017learning} and multimodal translation\cite{huang2018multimodal,DRIT_plus,kang2020unsupervised}. 

Exemplar-based image-to-image translation is a kind of image-to-image translation, which makes use of a structure image (input) and a style image (exemplar) to generate output. Most recent approaches for this task has achieved steady progress by leveraging deep CNNs due to its high capacity, but at the same time, requires a large-scale paired-domain dataset~\cite{isola2017image,zhang2020cross, park2019semantic, zhu2020sean,sangkloy2017scribbler}, thus having limited applicability. Among them, CoCosNet\cite{zhang2020cross} has been regarded as showing state-of-the-art results in exemplar-based image-to-image translation. 
It requires to train the networks, a large-scale task-specific domain benchmark, but in practice, collecting such a dataset is very costly and even impossible. 
Developing the technique to overcome this limitation and translate the image without intensive, exhaustive training on task-specific domains is our topic in this paper.

\begin{figure*}[t]
\begin{center}
\includegraphics[width=1\linewidth]{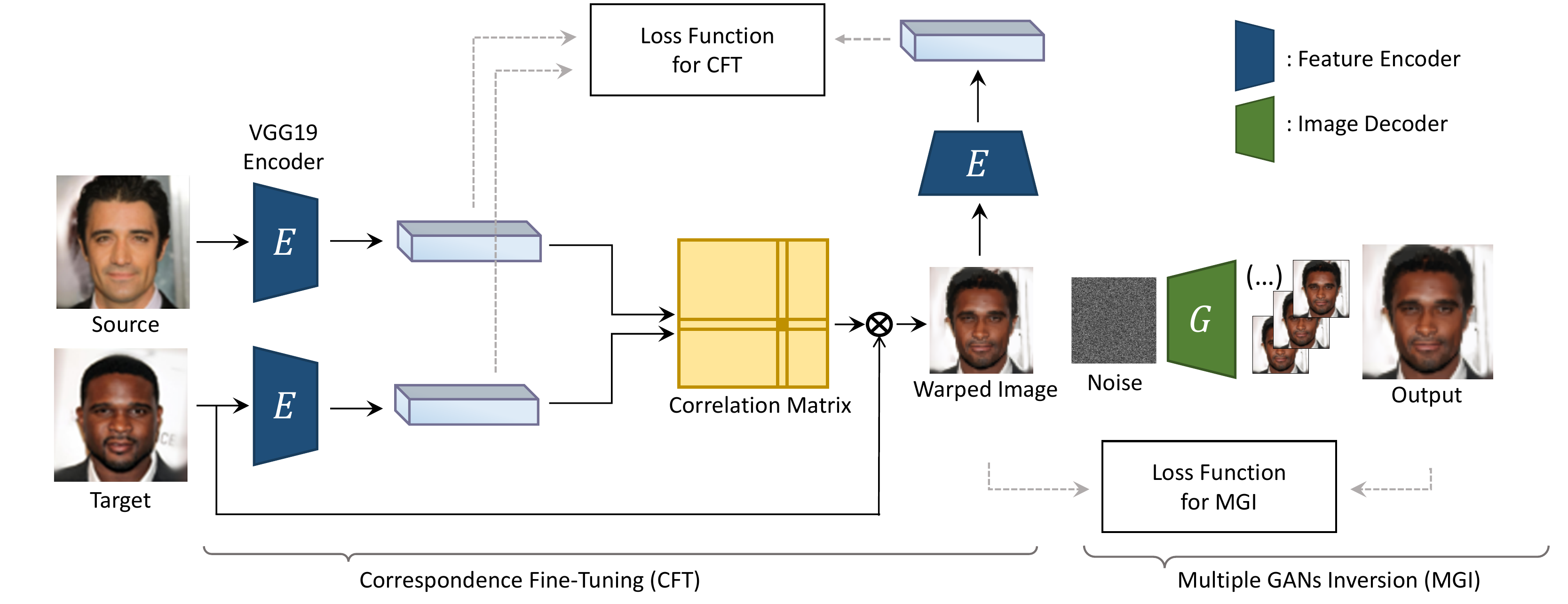}
\end{center}\vspace{-10pt}
  \caption{\textbf{Network configuration.} 
  Our architecture consists of two modules, namely correspondence fine-tuning and multi-GANs inversion. 
  At the first, we predict a translation hypothesis by first computing the similarity between each source point and all the target points and then warping the target image in a probabilistic manner. At the second, we refine the warped image to more natural and plausible one through the multi-GANs inversion.} \vspace{-10pt}
\label{cft_overall}
\end{figure*}

\textbf{Contrastive learning.}
Contrastive learning~\cite{chen2020simple,he2020momentum,wu2018unsupervised} minimizes a feature distance between the positive samples while maximizing feature distances between negative ones. It has been a effective way in solving tasks as unsupervised learning, such as visual representation learning~\cite{he2020momentum, chen2020simple, chen2020improved}, labeling~\cite{sermanet2018time}, captioning\cite{luo2018discriminability}, and image generation\cite{park2020contrastive}. 
The main issue in contrastive learning is how to set good positive and negative samples. We utilize this learning framework to optimize our networks.



\begin{figure}[t]
\begin{center}
\includegraphics[width=1\linewidth]{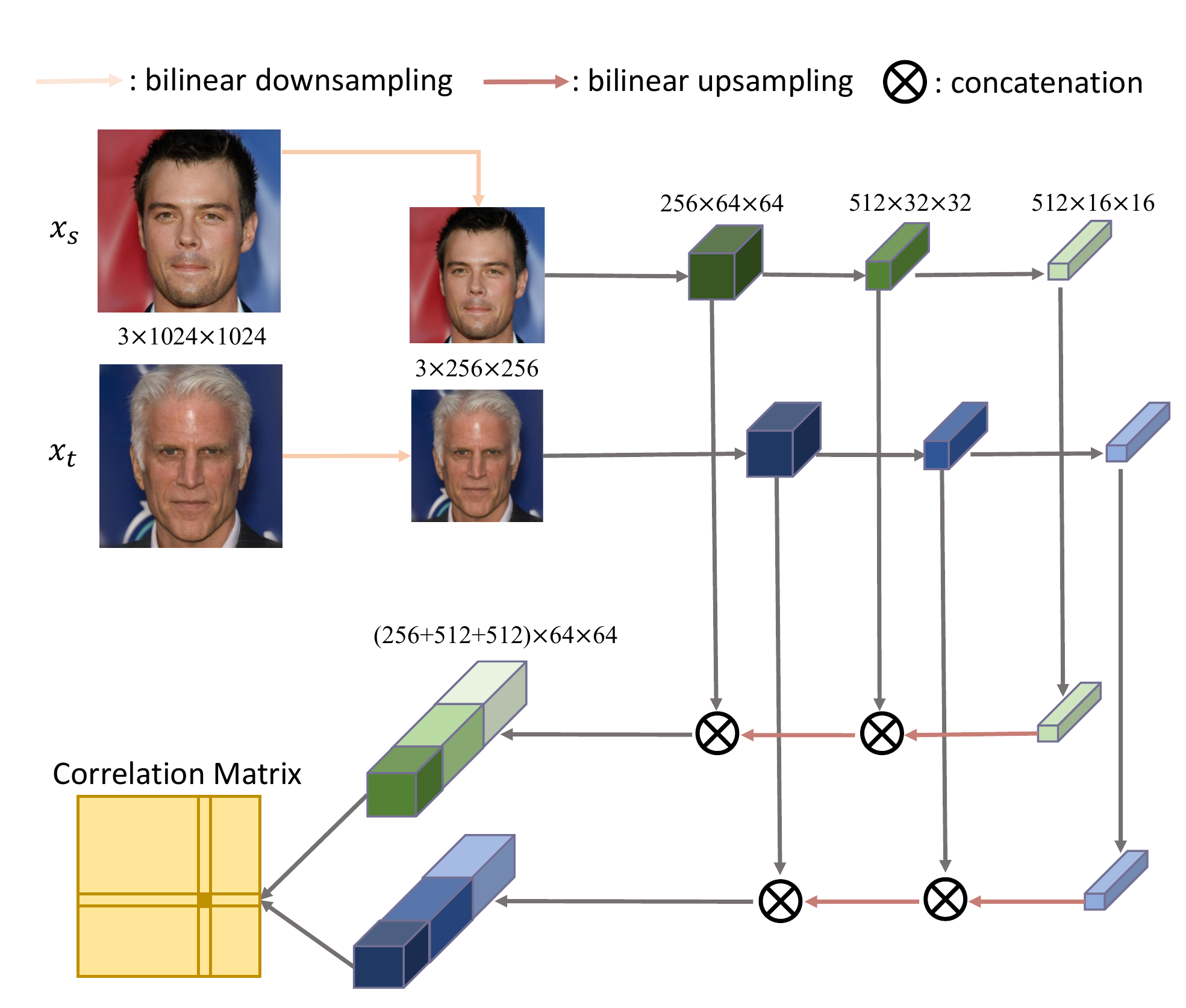}
\end{center}\vspace{-5pt}
  \caption{\textbf{Hierarchical feature encoding.} To measure the similarity in both low-level and high-level features, we extract multi-level features and concatenate them, followed by a bilinear sampler to upsample lower-level features, used to measure a correlation.}\vspace{-10pt}
\label{cft_detail}
\end{figure}

\textbf{GAN inversion}
GAN inversion aims at generating an image by solely optimizing a latent code of pre-trained GAN models, e.g., PGGAN~\cite{karras2017progressive} or StyleGAN~\cite{karras2019style}, given a target image, 
which has been popularly considered as an alternative to overcome the limitation of aforementioned task-specific GAN model recently. Traditional methods~\cite{jahanian2019steerability,menon2020pulse} make use of a single latent code, therefore have a limitation in representing diverse style. Recently, mGANprior\cite{gu2020image} provides to use multiple latent codes, but fails to predict optimized hyper-parameters in each image, which we solve in the paper.

\section{Methodology} 
\subsection{Motivation and Overview}
Exemplar-based image-to-image (I2I) translation aims at transferring the style of a target image $x_\mathcal{T}$ to the content of a source image $x_\mathcal{S}$, and generating an image $r_{\mathcal{T} \rightarrow \mathcal{S}}$. There have been many solutions to solve this, and recently most of the state-of-the-arts leverage deep CNNs to \emph{learn} the translation from source domain $\mathcal{S}$ to target domain $\mathcal{T}$ in which input source and target images are contained such that $x_\mathcal{S} \in \mathcal{S}$ and $x_\mathcal{T} \in \mathcal{T}$, respectively. They are generally trained on tremendous large-scale dataset of $\mathcal{S}$-$\mathcal{T}$ pairs for each specific task, e.g., edge-image or mask-image, to \emph{learn} mapping function, but collecting such dataset is challenging, costly and time-consuming, and in some scenario, it may be even impossible to collect such large-scale data. Training the model is another challenge. Indeed, most recent techniques were formulated in an adversarial learning~\cite{mao2017least,gulrajani2017improved,radford2015unsupervised}, and it has been very well-known that training this is notoriously difficult and requires enough time and memory. Limited generalization ability to unseen input image-pair is also limitation of existing solutions~\cite{zhu2017unpaired,park2020swapping,zhang2020cross,huang2018multimodal}.


In this paper, we instead explore an alternative, simple solution to solve exemplar-based image-to-image translation. Our key-ingredient is to solely utilize an input image pair $\{x_\mathcal{S},x_\mathcal{T}\}$, without depending on the exhausted training on task-specific, large-scale domain datasets $\{\mathcal{S},\mathcal{T}\}$, as in most previous approaches~\cite{zhang2020cross,park2019semantic,zhu2020sean}, to \emph{fine-tune} the mapping function between them from the off-the-shelf, general purpose, and task-agnostic feature and generator. 
In particular, rather than directly finding the mapping between them, we present a two-stage solution in a manner that on one hand, the model predicts a translation hypothesis by first computing the similarity between each source point and all the target points by means of the feature vectors, and on the other, the model refines for more plausible generation.

Specifically, at the first module, feature detector is used to extract the features $f_\mathcal{S}$ and $f_\mathcal{T}$ from an the input pair, $x_\mathcal{S}$ and $x_\mathcal{T}$, respectively, utilized to compute a correlation matrix. 
To improve the accuracy, deep convolutional neural networks may be a good candidate. To establish input image-specific correspondence fields, we fine-tune the networks through optimization with the proposed loss functions, including contrastive loss, contextual loss, and others (Sec.~\ref{sec:3_2}), by considering the matching statics within the input pair. At the second module, we refine the warped image to generate more natural and plausible image through the proposed multi-GANs inversion, using a self-deciding algorithm in choosing the hyper-parameters for GANs inversion (Sec.~\ref{sec:3_3}). In the following, we will introduce each module with its associated loss function. Figure~\ref{cft_overall} shows our pipeline of exemplar-based image-to-image translation.


\begin{figure}[t]
	\begin{center}
	\includegraphics[width=1\linewidth]{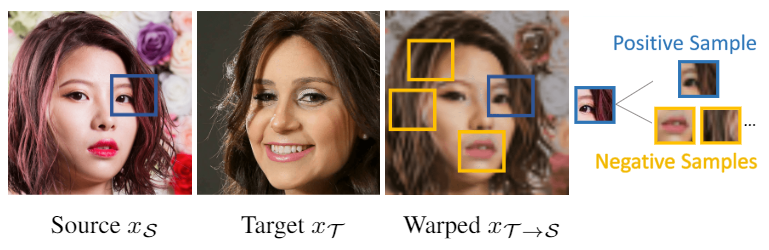}
	\end{center}
    \vspace{-10pt}
	\caption{\textbf{Examples of contrastive loss signal between source and target.} Since we cannot define the positive samples directly on the source and target images, we instead define the contrastive loss on source and warped images, based on the fact that the positive samples always occur at the same points across the images.}\vspace{-10pt}
	\label{cft_contrastive}
\end{figure}

\subsection{Correspondence Fine-Tuning}\label{sec:3_2}

In this section, we present correspondence fine-tuning(CFT) to measure the similarities between each point in source image $x_\mathcal{S}$ and all the other points in target image $x_\mathcal{T}$. 
It is inspired by the classical matching pipeline~\cite{rocco2017convolutional} in that we first extract the feature vectors and then compute the similarity between them. Such a architecture was also deployed in some other works, e.g., CoCosNet~\cite{zhang2020cross}, but they require the intensive, exhausted training on task-specific domain benchmarks $\{\mathcal{S},\mathcal{T}\}$, while our approach does not. 

\begin{figure}[t]
	\begin{center}
	\includegraphics[width=1\linewidth]{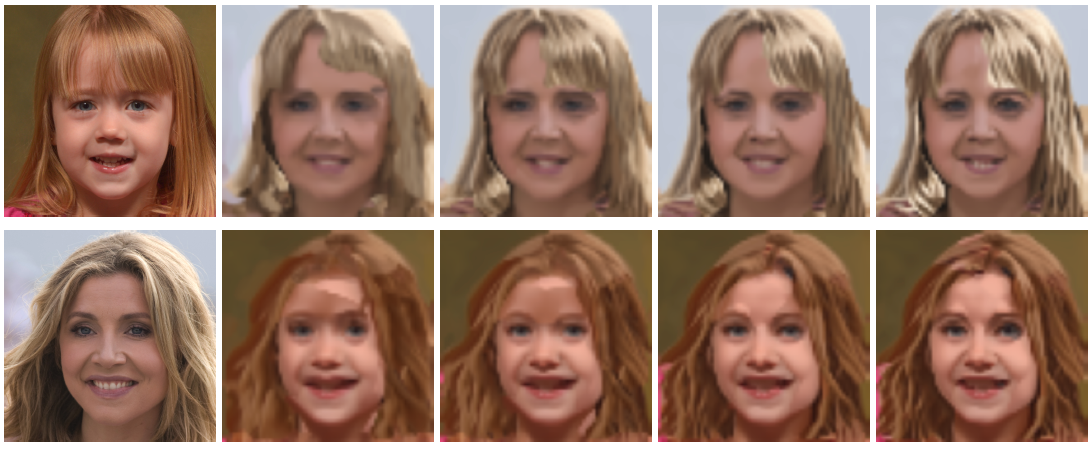}
    \vspace{-10pt}
	\caption{\textbf{CFT optimization.} (from left) Source and target as input, and following warped images as number of iteration increases, shows how our technique converges to well-established correspondence via optimization process.}\vspace{-10pt}
	\end{center}
	\label{cft_optimization}
\end{figure}

\textbf{Architecture.} Specifically, we first extract the deep convolutional feature, e.g., VGGNet~\cite{simonyan2014very} or ResNet~\cite{he2016deep} pretrained on ImageNet~\cite{russakovsky2015imagenet}, as follows:
\begin{align}
    f_\mathcal{S} &= F_\mathcal{S}(x_\mathcal{S};\mathbf{W}_\mathcal{S}) \in \mathbb{R}^{H\times W\times C}, \\
    f_\mathcal{T} &= F_\mathcal{T}(x_\mathcal{T};\mathbf{W}_\mathcal{T}) \in \mathbb{R}^{H\times W\times C}, 
\end{align}
where $F(x;\mathbf{W})$ denotes a feed-forward process of input $x$ through a deep network of parameters $\mathbf{W}$, and $H$ and $W$ are spatial size of the feature, with $C$ channels. $\mathbf{W}_\mathcal{S}$ and $\mathbf{W}_\mathcal{T}$ are network parameters for encoding source and target features, respectively. 
To measure the similarity in both low-level and high-level features, we extract multi-level features and concatenate them, followed by a bilinear sampler to upsample lower-level features to the highest-level feature, used to measure a correlation matrix. Figure~\ref{cft_detail} illustrate the procedure to extract such a hierarchical feature. We then compute a correlation matrix $\mathcal{M} \in \mathbb{R}^{HW \times HW}$ of which each term is a pairwise feature correlation as
\begin{equation}
    \mathcal{M}(u,v) = \frac{\hat{f}_\mathcal{S}(u)^T \hat{f}_\mathcal{T}(v)}{|\hat{f}_\mathcal{S}(u)|\cdot|\hat{f}_\mathcal{T}(v)|},
\end{equation}
where $u$ and $v$ represent all the points in the source and target images, respectively. $\hat{f}_\mathcal{S}(u)$ and $\hat{f}_\mathcal{T}(v)$ are the channel-wise centralized feature of $f_\mathcal{S}$ and $f_\mathcal{T}$ at the position u and v as follows:
\begin{align}
    \hat{f}_\mathcal{S}(u) &= f_\mathcal{S}(u) - \text{mean}(f_\mathcal{S}), \\
    \hat{f}_\mathcal{T}(v) &= f_\mathcal{T}(v) - \text{mean}(f_\mathcal{T}), 
\end{align}
where $\text{mean}(f_\mathcal{S})$ is an average of $f_\mathcal{S}(u)$ for all $u$ across all the points in the source. $\text{mean}(f_\mathcal{T})$ is similarly defined.
Since $M(u,v)$ represents a similarity between points $u$ and $v$, the higher, the more similar.

By using the similarity $\mathcal{M}$, we reconstruct the warped target image $r_{\mathcal{T} \rightarrow \mathcal{S}} = \psi(x_\mathcal{T},\mathcal{M})\in \mathbb{R}^{H\times W}$. The function $\psi(\cdot)$ could be formulated in many possible ways, but we borrow the technique as in~\cite{zhang2020cross} that uses a reconstruction formulation:
\begin{equation}
    r_{\mathcal{T} \rightarrow \mathcal{S}}(u) = \sum_{v}^{} \underset{v}{\mathrm{softmax}} (\alpha \mathcal{M}(u,v)) \cdot x_\mathcal{T}(v), 
\end{equation}
where $\underset{v}{\mathrm{softmax}}$ means the softmax operator across $v$.

\begin{figure}[t]
	\begin{center}
	\includegraphics[width=1\linewidth]{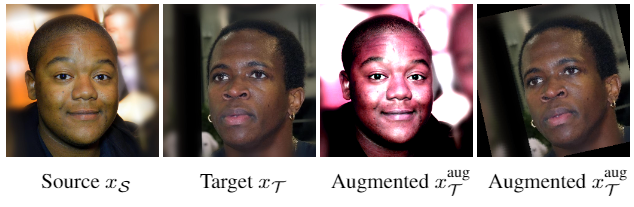}
    \end{center}
    \vspace{-10pt}
	\caption{\textbf{Examples of augmented source and target images.} To overcome the restriction that accesses original source and target images during optimization in our framework, we deploy an online augmentation in a manner independent augmentations are applied to source and target images at each iteration.}\vspace{-10pt}
	\label{cft_augmentation_figure}
\end{figure}

\textbf{Loss function.}
In general, to warp the target image $x_\mathcal{T}$ to reference accurately, the \emph{well-trained} network parameters $\mathbf{W}_\mathcal{S}$ and $\mathbf{W}_\mathcal{T}$ are required that are capably of extracting reliable both low-level and high-level information. Conventionally, they are learned on large-scale training dataset of $\{\mathcal{S},\mathcal{T}\}$, but in our setting, we only have an input image $x_\mathcal{S}$ and $x_\mathcal{T}$. Inspired by~\cite{ulyanov2018deep, shaham2019singan}, we \emph{fine-tune} the parameters $\mathbf{W}_\mathcal{S}$ and $\mathbf{W}_\mathcal{T}$, from the off-the-shelf feature detector such as VGG~\cite{simonyan2014very} or ResNet~\cite{he2016deep} that are not tailored to specific matching tasks, but developed for general purpose, to generate accurate correspondence fields specifically for the input pair. Desirable properties of the optimized feature are that they should extract the features are similar to each other for the matching points across source and target, dissimilar otherwise. The contrastive learning~\cite{chen2020simple,park2020contrastive} might be a possible solution to learn such a discriminative feature descriptor, but we do not have the positive and negative pairs to define the contrasitive loss function.

Alternatively, by definition, we can set the \emph{pseudo} positive samples between $x_\mathcal{S}$ and 
$r_{\mathcal{T} \rightarrow \mathcal{S}}$. 
Specifically, we use an InfoNCE loss~\cite{oord2018representation,he2020momentum} for contrastive learning between $f_\mathcal{S}$ and $f_{\mathcal{T} \rightarrow \mathcal{S}} = F(r_{\mathcal{T} \rightarrow \mathcal{S}};\mathbf{W}_\mathcal{T})$, defined such that
\begin{equation}
\begin{split}
    &l(f, f^{+}, f^{-}) \\
    &= -\mathrm{log} [ \frac{\exp (f \cdot f^+ / \tau )}{\exp (f \cdot f^+ / \tau) + \sum^{N}_{n=1} \exp(f \cdot f^{-}_n / \tau ) } ],
\end{split}
\end{equation}
where $f$ and $f^+$ are positive samples, namely $f_\mathcal{S}$ and $f_{\mathcal{T} \rightarrow \mathcal{S}}$, while $f$ and $f^-$ are negative samples. We then define our constrastive loss $\mathcal{L}_{cont}$, which helps to match that input-output patches at a specific location, defined as follows:
\begin{equation}
    \mathcal{L}_{cont}
    = \sum_{u} \sum_{n \in \mathcal{N}_{u}} l(f_\mathcal{S}(u), f_{\mathcal{T} \rightarrow \mathcal{S}}(u), f_{\mathcal{T} \rightarrow \mathcal{S}}(n)),
\end{equation}
where $\mathcal{N}_{u}$ is the negative samples. In our method, we set all other points except for point $u$ as the negatives. 
Thanks to this loss function of parameters $\mathbf{W}$, we can fine-tune the feature extraction networks without accessing ground-truth correspondence fields. Figure~\ref{cft_contrastive} illustrates the procedure to define positive and negative in the contrastive learning.

\begin{figure}[t]
\begin{center}
\includegraphics[width=1\linewidth]{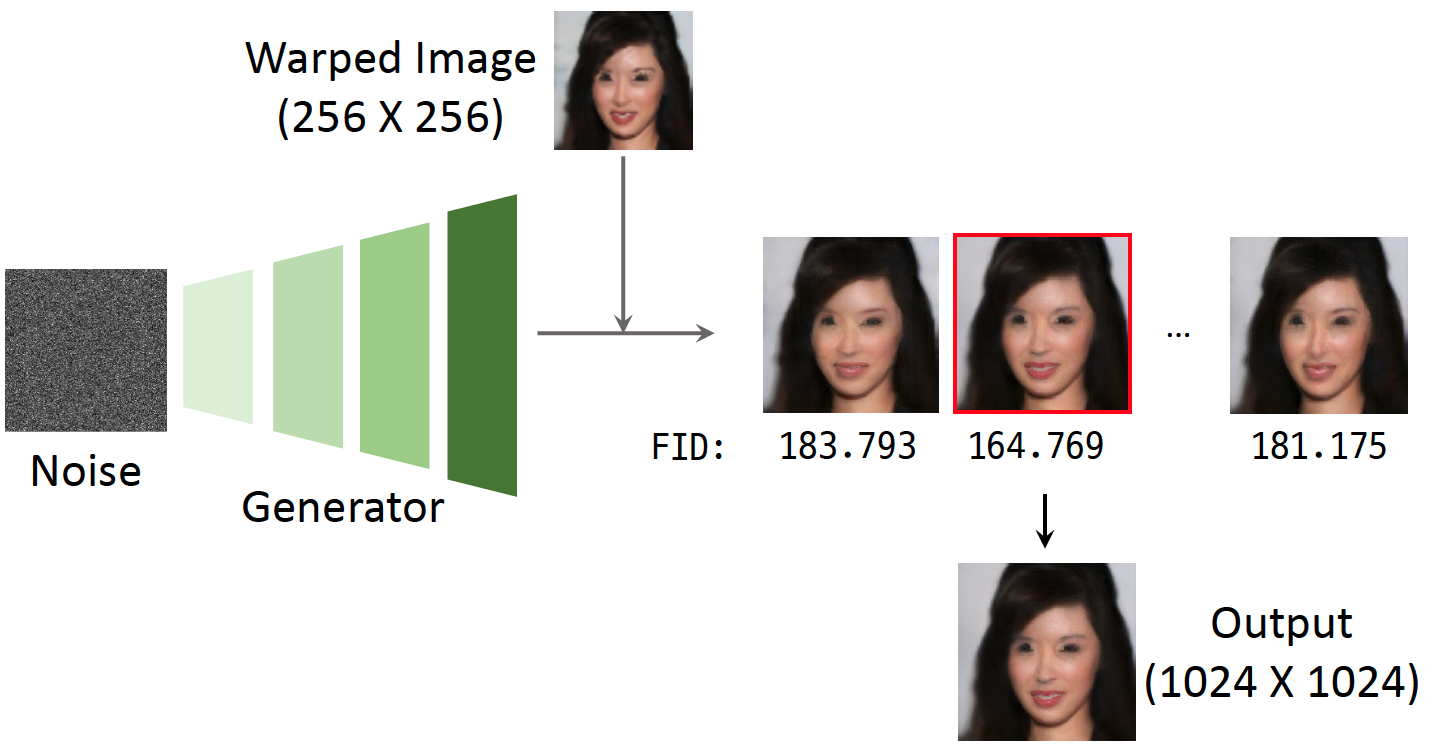}
\end{center}
\vspace{-10pt}
  \caption{\textbf{Illustration of multi-GANs inversion.} Our multiple GAN inversion network generates high-resolution image from warped image and this model can generate high-quality results from low-resolution warped image. Multiple GAN Inversion also includes innovative self-deciding algorithm in choosing the hyperparameters using FID that avoids human intervention.}\vspace{-10pt}
\label{mgan_architecture}
\end{figure}

In addition, while utilizing $\mathcal{L}_{cont}$ helps to reconstruct the translation image $x_{\mathcal{T} \rightarrow \mathcal{S}}$ having a content structure similar to the source image $x_{\mathcal{S}}$, but may lose some style information of target image $x_{\mathcal{T}}$ without additional losses to preserve this. 
To overcome this, we additionally use two more loss functions, namely perceptual loss and context loss. First of all, we utilize the perceptual loss to reduce semantic discrepancy between target and warped image as 
\begin{equation}
\begin{split}
    &l_{perc}^{\phi} \left(x_T,x_{T \rightarrow S}\right) \\
    &= \sum_{c,c'}
    \|G^{\varnothing}(x_\mathcal{T})(c,c') - G^{\varnothing} (x_{\mathcal{T} \rightarrow \mathcal{S}}(c,c')) \|^2 _2, \\
    &G^{\varnothing}(c,c') = \sum_{u} \varnothing(u,c) \varnothing(u,c'),
    \end{split}
\end{equation}
where $\varnothing(\cdot)(u,c)$ represents the activation by pre-trained VGG encoder of the $c$-th filter at position $u$ and $G^{\varnothing}$ is the Gram matrix~\cite{johnson2016perceptual}. 
Additionally, we use a contextual loss function, defined as follows:
\begin{equation}
\begin{split}
    &\mathcal{L}_{context} \\
    &= -\text{log} \left( \sum_{u} \max_l \mathcal{X}\left(\varnothing (x_\mathcal{T})(u), \varnothing (x_\mathcal{\mathcal{T} \rightarrow \mathcal{S}}\right)(l)\right),
\end{split}
\end{equation}
where $\mathcal{X}(\cdot,\cdot)$ is the contextual similarity function defined in~\cite{mechrez2018contextual}.


Similarly to~\cite{ulyanov2018deep, shaham2019singan}, we optimize the total loss function 
\begin{equation}
    \mathcal{L}_{total} = \mathcal{L}_{cont} + \lambda_{perc} \mathcal{L}_{perc} + \lambda_{context} \mathcal{L}_{context}
\end{equation}


\begin{figure}[t]
	\begin{center}
	\includegraphics[width=1\linewidth]{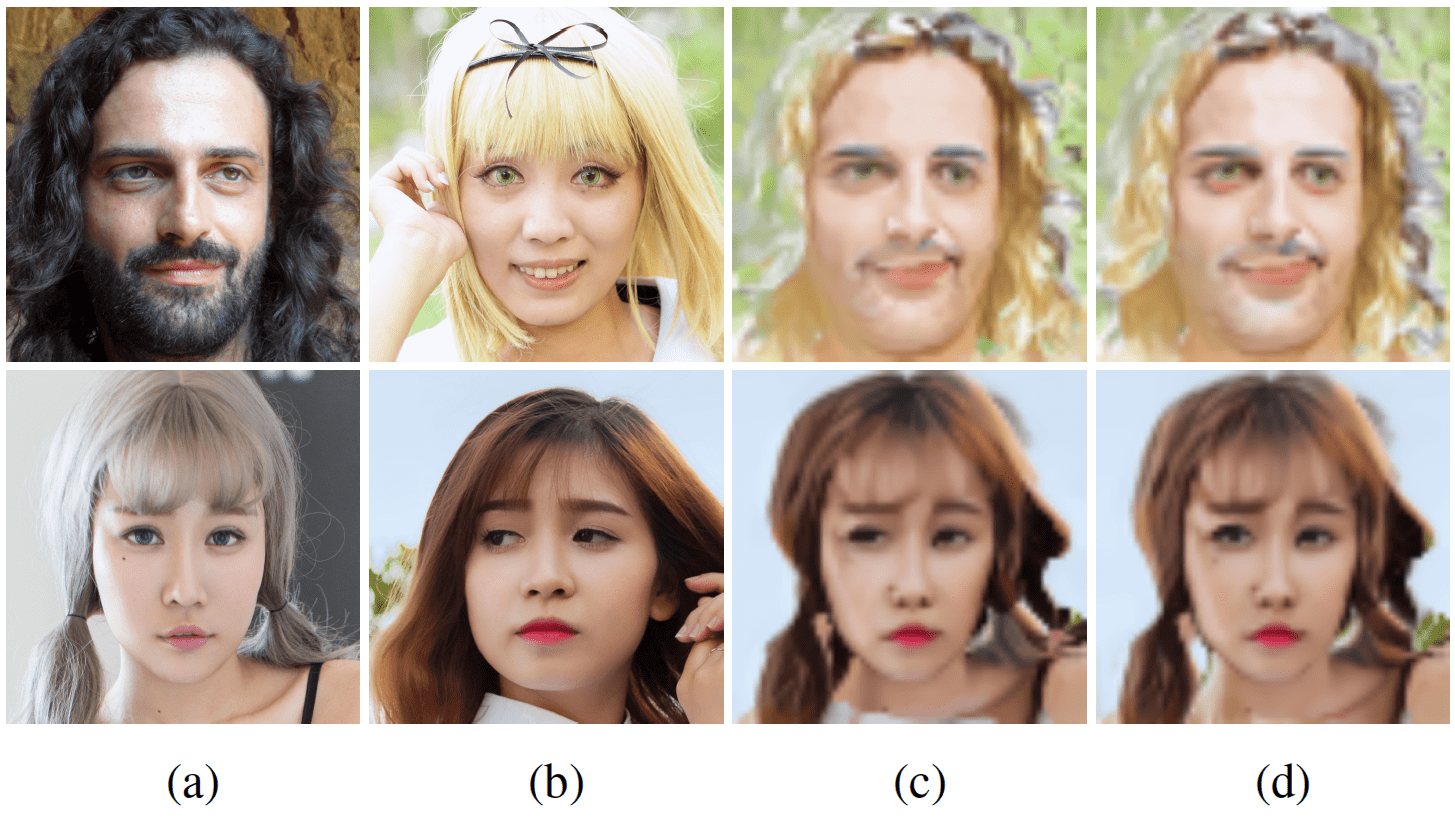}
    \end{center}
    \vspace{-10pt}
	\caption{\textbf{Effectiveness of online augmentation.} (a) source, (b) target, warped images (c) without and (d) with online augmentation. Compared to (c), (d) is better representing structure and style, showing that augmentation makes warped image more realistic.
	}\vspace{-10pt}
	\label{oeft_augmentation}
\end{figure}

by the ADAM optimizer, where $\lambda$ is a hyper-parameter to be tuned. Note the any optimizer to minimize the $\mathcal{L}_{total}$ can be used. As shown in Figure~\ref{cft_optimization}, our method gradually converges to establish the optimal correspondences across source and target images as iteration process is proceeded, without any prior training on the domain or ground-truth supervision.

\textbf{Online adaptation during optimization.} 
To overcome the strict restriction in our setting that accesses only source $x_{\mathcal{S}}$ and target $x_{\mathcal{T}}$ images during optimization, we deploy an online augmentation in a manner independent augmentations are applied to source and target images at each iteration. Specifically, we apply photometric deformations, such as color jittering or gaussian noise, to source $x_{\mathcal{S}}$ and geometric deformations, such as scaling or affine transformation, to target $x_{\mathcal{T}}$, and use the augmented source $x^\text{aug}_{\mathcal{S}}$ and target $x^\text{aug}_{\mathcal{T}}$ dynamically during optimization, as exemplified in Figure \ref{cft_augmentation_figure}, where we find out that such online augmentation during optimization dramatically improves the quality of fine-tuned features.

\subsection{Multiple GANs Inversion}\label{sec:3_3}
Even though the aforementioned technique generates the translation image, the output inherently has a limited resolution, due to the memory restriction, and some artifacts. To overcome these, CoCosNet~\cite{zhang2020cross} attempts to train additional translation networks, but it requires additional burden to train. 
Based on our design philosophy that only uses an off-the-shelf, general purpose network, we also present a method to utilize the GAN inversion~\cite{gu2020image} in that we only optimize the latent code that is likely to generate the plausible image guided by the warped image $r_{\mathcal{T} \rightarrow \mathcal{S}}$ with the pre-trained, fixed generation network of parameters $\mathbf{W}_{I}$. 
Even though any GAN inversion technique can be used, in this paper, we use recent one, namely mGANprior~\cite{gu2020image}. 
To further improve the performance of this, we present a self-deciding algorithm when choosing the hyper-parameters of GAN inversion based on Fréchet Inception Distance (FID)~\cite{heusel2017gans}, which is a no-reference quality measure for image generation, so we do not need any supervision.

Specifically, we reformulate the GAN inversion module to generate multiple hypotheses using different number of layers. Among the multiple hypotheses $\{y_1, ..., y_N\}$ with the number of hypotheses $N$, we decide the most plausible one based on FID scores, which enables to make the inversion result be in natural data distributions. We define the latent code for $n$-th attempt as $z^n_I$, which can be found by minimizing the distance function between $\text{down}(y_n)$ and $r_{\mathcal{T} \rightarrow \mathcal{S}}$ as 
\begin{equation}
    z^n_{I} = \text{argmin}_{z} {\mathcal{D}(\text{down}(y_n), r_{\mathcal{T} \rightarrow \mathcal{S}};W^n_{I})},
\end{equation}
where $\text{down}(\cdot)$ is the downsampling operator, and $y_n=F(z;W^n_{I})$. $\mathbf{W}_{I}$ is an inversion network parameter, so $\mathbf{W}^n_{I}$ is the parameters of $n$-th attempt. $\mathcal{D}(\cdot,\cdot)$ is the distance function, which can be L1, L2, or perceptual loss function~\cite{johnson2016perceptual}. 
By measuring the FID scores of reconstructed images such that $k_n$, and finding the minimum, we final get the final translation image $g = y_{n^*}$ such that $n^* = \text{min}_{n} {k_n}$. Figure~\ref{mgan_architecture} visualizes our aforementioned procedure.



\section{Experiments}
In this section, we report all the details including implementation and training details to reproduce our experiments.

\begin{figure}[t]
\begin{center}
\includegraphics[width=1\linewidth]{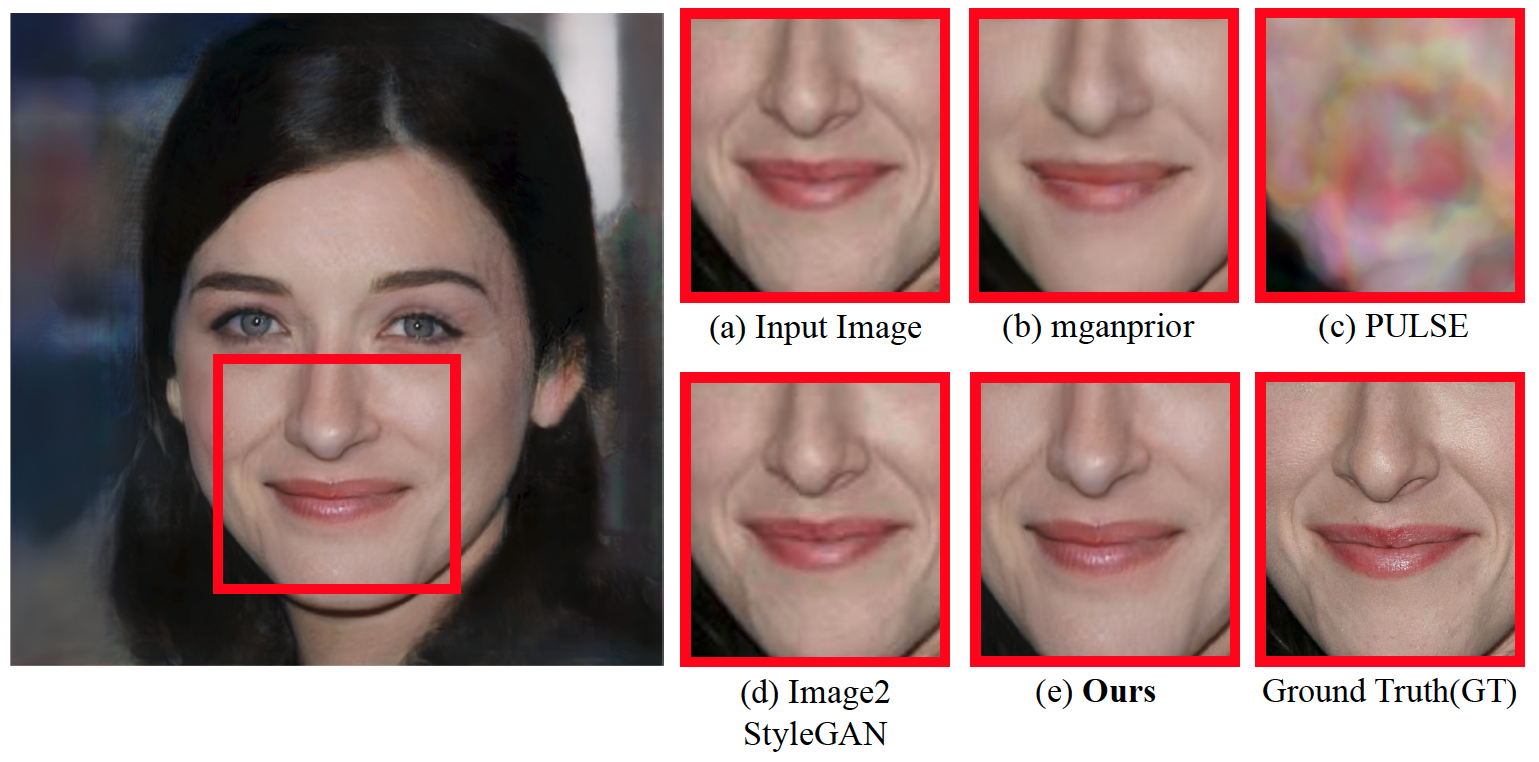}
\end{center}
\vspace{-10pt}
  \caption{\textbf{Comparison of GANs Inversion result with existing methods on CelebA-HQ dataset.} Our model generates competitive results with the smallest computational cost.} 
\label{gan_inversion_celebahq}
\end{figure}

\begin{figure*}[t]
	\centering
	\includegraphics[width=1\linewidth]{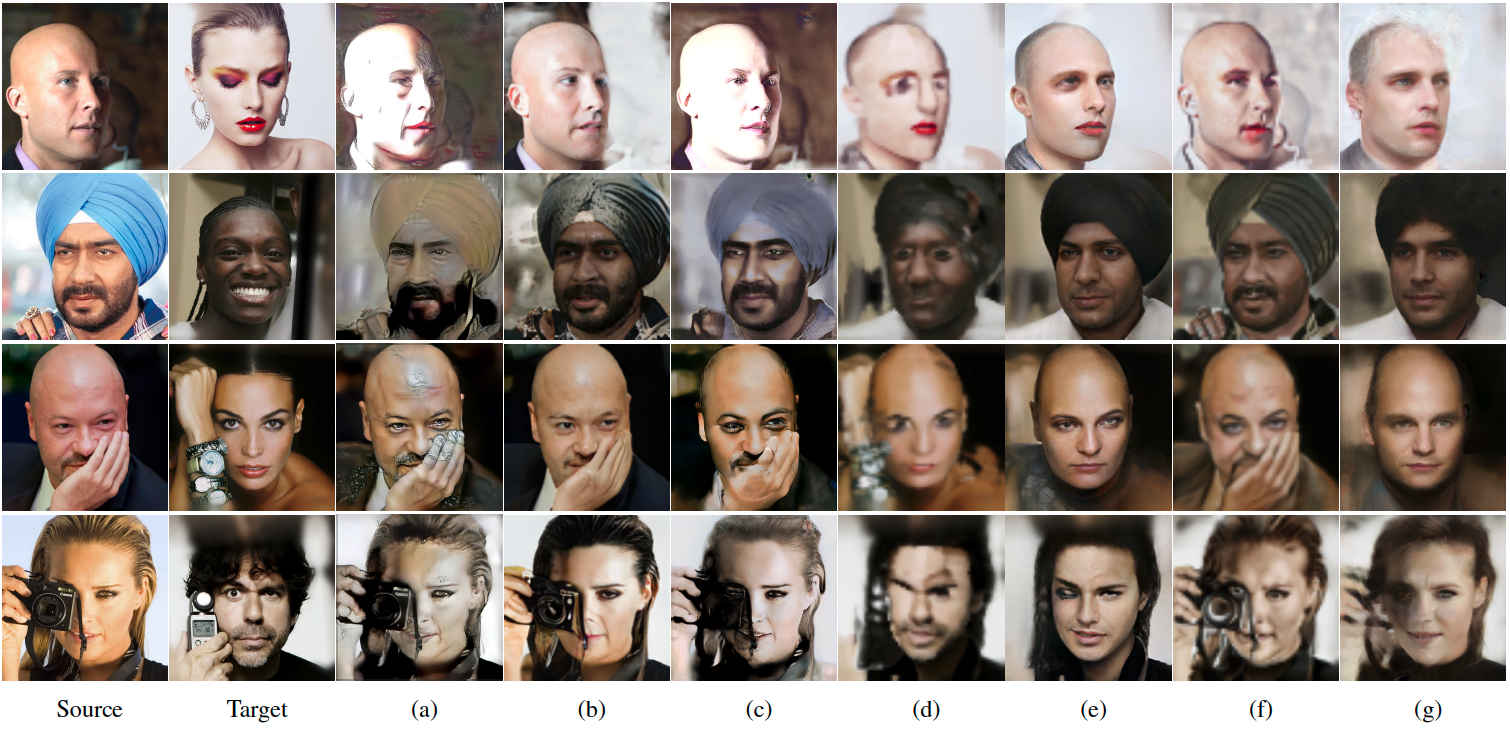}
	\caption{\textbf{Comparison of OEFT with other existing exemplar-based image-to-image translation and style transfer methods on CelebA-HQ dataset.} Given source and target images, translation results by (a) Style Transfer~\cite{gatys2015neural}, (b) Swapping Autoencoder~\cite{park2020swapping}, (c) Image2StyleGAN~\cite{abdal2019image2stylegan}, (d) CoCosNet~warped~\cite{zhang2020cross}, (e) CoCosNet-final~\cite{zhang2020cross}, (f) Ours-warped, (g) Ours-fin. This shows that our networks can translate local features as well as global features from both structure and style. Note that (d), (e) are trained on CelebA-HQ mask2face dataset, including supervised segmentation mask on CelebA-HQ.}\vspace{-10pt}\label{celebahq_result}
\end{figure*}

\begin{figure}[t]
\begin{center}
\includegraphics[width=1\linewidth]{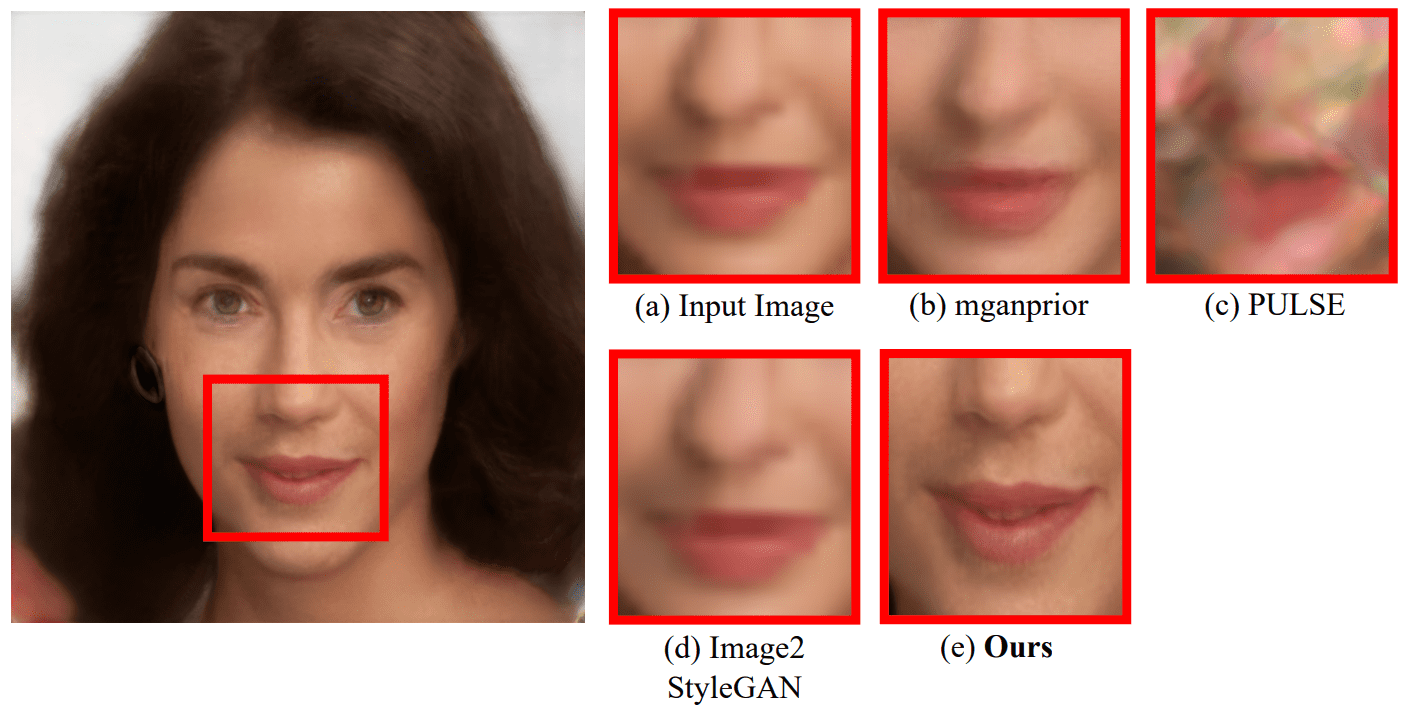}
\end{center}
\vspace{-10pt}
  \caption{\textbf{Comparison of GANs Inversion result with existing methods on CoCosNet warped image.} Our model generates competitive results with the smallest computational cost.}\vspace{-10pt}
\label{gan_inversion_warped}
\end{figure}

\subsection{Implementation Details}
We first summarize implementation details of our approach, especially in correspondence fine-tuning and multiple GANs inversion. For CFT module, we optimized our networks over 200 iterations inside the CFT network. A pair of source and target is bilinear down-sampled to the size of 256$\times$256. This pair would be augmented at every iteration. The target should be augmented by Geometric augmentation, the source should be augmented by Photometric augmentation. During each iteration, both the augmentation-pair and the original pair start passing through the network, in order to improve stable learnability.


For MGI module, we basically try to follow the default setting of mGANprior, with marginal modifications. We hypothesize composing layers ranging to 4 to 8, Number of the latent codes ranging to 10 to 40. Up-sampling factor was 4 for processing 256 $\times$ 256 to 1024 $\times$ 1024 image. We used PGGAN-Multi-Z and StyleGAN for mGANprior. For the distance function, we used L2 and perceptual Loss, as in mGANprior. We used a batch size of 1.

\subsection{Experimental Setup}
\textbf{Datasets.} We used two kinds of datasets to evaluate our method, namely CelebA-HQ\cite{liu2015deep} and Flickr Faces HQ(FFHQ)\cite{karras2019style}. During the optimization, where input images were resized into 256 $\times$ 256.

\textbf{Baselines.} 
We compared our method with recent state-of-the-art exemplar-based image-to-image translation methods such as CoCosNet~\cite{zhang2020cross}, Swapping Autoencoder~\cite{park2020swapping}, Image2StyleGAN\cite{abdal2019image2stylegan}, and Style Transfer\cite{gatys2015neural}.
It should be emphasized all the methods were trained on tremendous training samples, while our method just fine-tune the off-the-shelf networks without training for image-to-image translation. In addition, to evaluate our correspondence fine-tuning module, warped image output is compared with that of CoCosNet~\cite{zhang2020cross}. Finally, we also evaluate our GAN inversion module in comparison with mGANprior~\cite{gu2020image}, PULSE~\cite{menon2020pulse}, and Image2StyleGAN~\cite{abdal2019image2stylegan}, which have been state-of-the-art in GAN inversion. 

\begin{figure*}[t]
	\centering
	\includegraphics[width=1\linewidth]{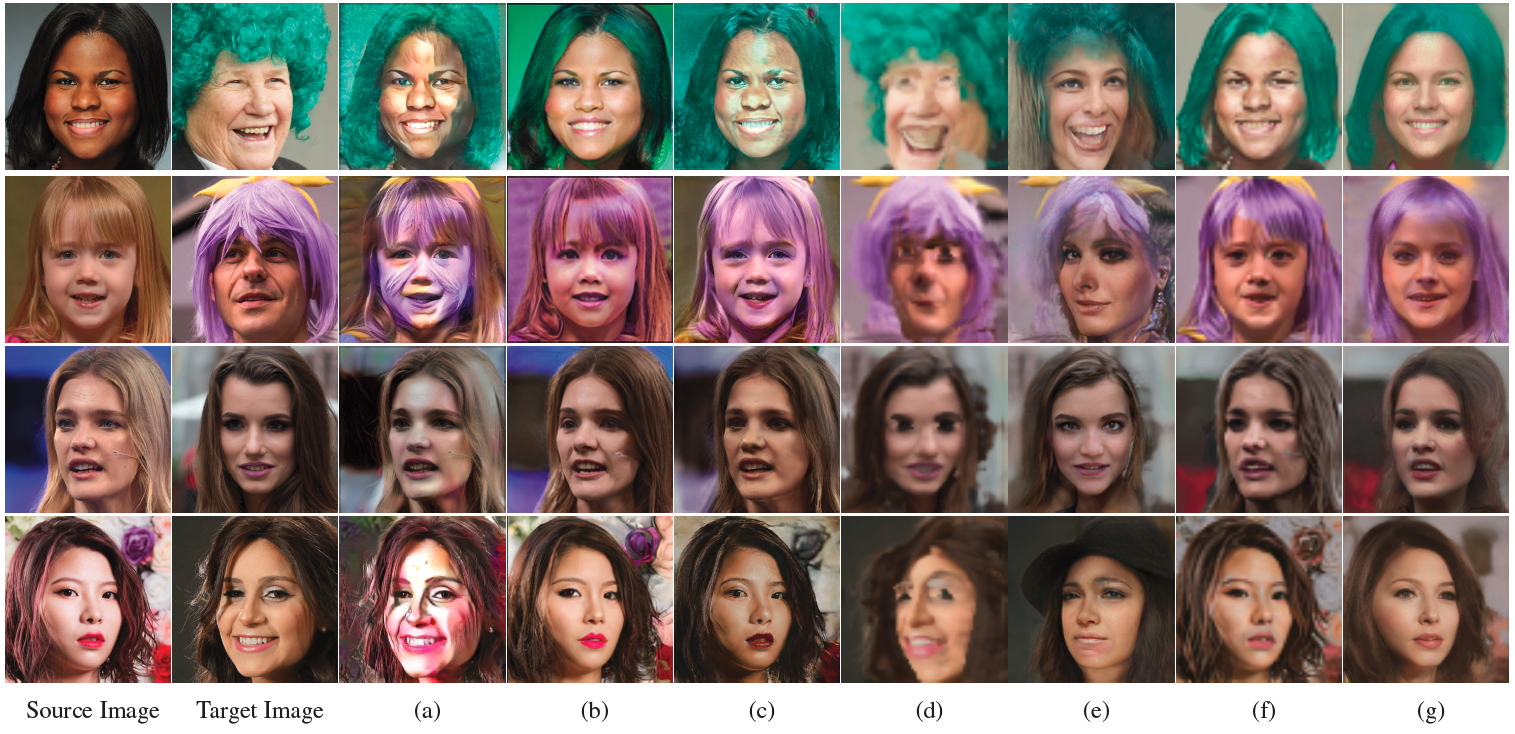}
	\caption{\textbf{Comparison of OEFT with other existing exemplar-based image-to-image translation and style transfer methods on FFHQ dataset.} Given source and target images, translation results by (a) Style Transfer~\cite{gatys2015neural}, (b) Swapping Autoencoder~\cite{park2020swapping}, (c) Image2StyleGAN~\cite{abdal2019image2stylegan}, (d) CoCosNet~warped~\cite{zhang2020cross}, (e) CoCosNet-final~\cite{zhang2020cross}, (f) Ours-warped, (g) Ours-fin. This shows that our networks can translate local features as well as global features from both structure and style, and generates high-resolution image. Note that (b), (c) are trained on FFHQ dataset.}\vspace{-10pt} \label{ffhq_result}
\end{figure*}

\subsection{Experimental Results}
Since our framework consists of two sub-networks, CFT and MGI, in this section, we evaluate the two modules separately, specifically \emph{ours-warp} denotes the results by only CFT module, and \emph{MGI} denotes the inversion results by only MGI module. We evaluate our final results as \emph{ours-final}.

\textbf{Qualitative evaluation.} We first evaluate the effectiveness of online augmentation in the CFT module in Figure~\ref{oeft_augmentation}, which shows that by employing online augmentation, we can more effectively warp the style of target image. We also evaluate our MGI module in Figure~\ref{gan_inversion_celebahq}, where we show that our generated results are more realistic and plausible. For instance, our method better reflects eyes or wrinkle as well as overall structure, clearly outperforming the state-of-the-art methods. 
In addition, our approach has shown the high generalization ability to any unseen input images, which the other previous methods often fail.

\begin{table}
\begin{center}
\begin{tabular}{|l|c|c|}
\hline
Model & celebahq & FFHQ \\
\hline\hline
Style Transfer & 296.064 & 327.404 \\
Swapping Autoencoder & 347.284 & 377.285 \\
Image2StyleGAN & 292.070 & 332.984 \\
CoCosNet(warp) & 391.688 & 433.389 \\
CoCosNet(result) & 298.185 & 323.914 \\
Ours(warp) & 346.065 & 394.312 \\
Ours(result) & \textbf{232.726} & \textbf{302.048} \\

\hline
\end{tabular}
\end{center}
\vspace{-5pt}
\caption{\textbf{Quantitative Evaluation of our OEFT.} We used Fréchet Inception Distance(FID) to measure the performance of various network structures.}\vspace{-10pt}
\label{quantitative_cft}
\end{table}

More qualitative results are shown in Figure~\ref{celebahq_result} and Figure~\ref{ffhq_result}. Our generated results are realistic and represent both global-local visual features while successfully preserving the structure from source. Our success in terms of fine style information, in specific, can be detailed seen in Figure~\ref{gan_inversion_celebahq}. For example, lip color and lip crease as well as overall clarity outperform the state-of-the-arts. Furthermore, since OEFT is not task-specific and developed for more generalization, it can work for a number of new combinations of tasks. The examples shown here are only a part of them. 

We further evaluate OEFT network in Table~\ref{quantitative_cft}, Table~\ref{quantitative_mgan} on CelebA-HQ(MGAN, OEFT CelebA-HQ result), FFHQ dataset(OEFT FFHQ result), showing the OEFT evaluation results under the FID metric. Our model significantly outperforms the state-of-the-arts under the evaluation metrics. 

\begin{figure}[t]
\begin{center}
\includegraphics[width=0.95\linewidth]{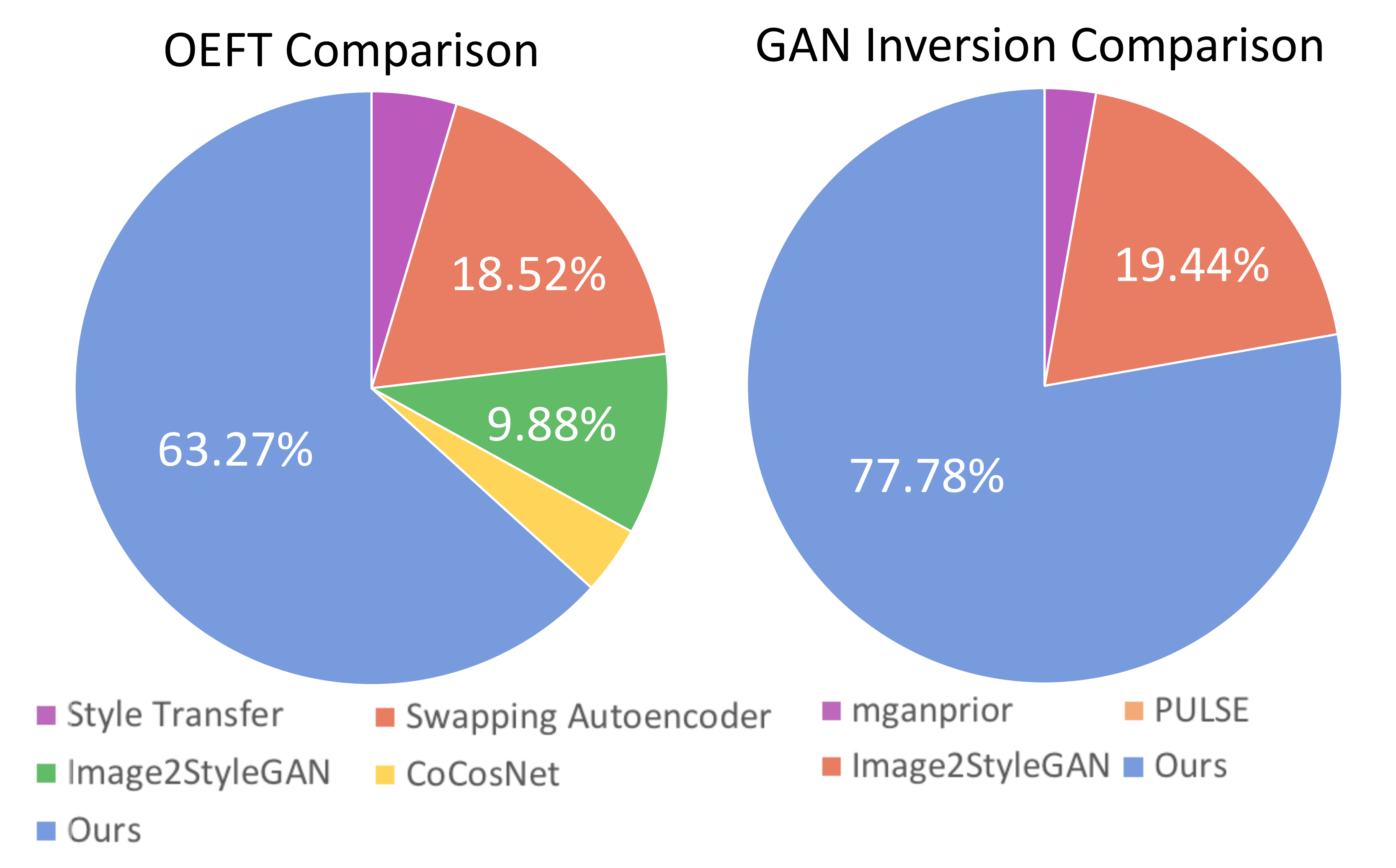}
\end{center}\vspace{-15pt}
  \caption{User Study result.}
\label{fig_userstudy}\vspace{-10pt}
\end{figure}

\textbf{User study.} We also conducted a user study on 80 participants. Figure~\ref{fig_userstudy} shows our user-study result. In OEFT comparison, there are 63.27\% users that prefer image quality for our method. In FFHQ, 77.78\% users prefer our image quality and most respondent prefers structure of source and style of target.

\begin{table}
\begin{center}
\begin{tabular}{|l|c|c|}
\hline
Model & celebahq & warped \\
\hline\hline
mganprior & 74.749 & 194.780 \\
PULSE & 449.530 & 427.773 \\
Image2StyleGAN & 71.114 & 311.077 \\
Ours & \textbf{50.427} & \textbf{134.706} \\
\hline
\end{tabular}
\end{center}
\vspace{-5pt}
\caption{\textbf{Quantitative Evaluation our GAN Inversion.} We used Fréchet Inception Distance(FID) to measure the performance of various network structures.}
\label{quantitative_mgan}\vspace{-10pt}
\end{table}

\section{Conclusion}
In this paper, we proposed, for the first time, a novel framework, called online exemplar fine-tuning (OEFT), to fine-tune the pre-trained general-purpose networks for exemplar-based image-to-image translation task, without need of any large-scale dataset, hand-labeling, and task-specific training process. 
We formulate overall networks as two sub-networks, including correspondence fine-tuing (CFT) and multiple GANs inversion (MGI). 
By optimizing the feature extractor in the CFT and the latent codes for image generator in the MGI to an input image pair, our approach achieves high generalization ability to unseen image pairs, which has been one of the major bottleneck of previous methods. Experimental results on a variety of benchmarks and in comparison to state-of-the-arts that based on learning on domain- or tasks-specific dataset proved that our method even outperforms the existing solutions.

{\small
\bibliographystyle{ieee_fullname}
\bibliography{cvpr}
}

\end{document}